\documentclass[conference]{IEEEtran}
\IEEEoverridecommandlockouts
\usepackage{cite}
\usepackage{amsmath,amssymb,amsfonts}
\usepackage{algorithmic}

\usepackage{pgfplotstable}
\usepackage{pgfplots}
\usepackage{graphicx}
\usepackage[ruled,vlined]{algorithm2e}
\usepackage{textcomp}
\def\BibTeX{{\rm B\kern-.05em{\sc i\kern-.025em b}\kern-.08em
    T\kern-.1667em\lower.7ex\hbox{E}\kern-.125emX}}
\begin{document}

\title{Deep Learning Model Explainability for Inspection Accuracy Improvement in the Automotive Industry\\

\thanks{Supported by Faurecia}
}

\author{\IEEEauthorblockN{Anass EL HOUD}
\IEEEauthorblockA{\textit{Faurecia Clean Mobility} \\
\textit{Ecole Centrale de Lyon}\\
Bavans, France \\
anass.elhoud@faurecia.com}
\and
\IEEEauthorblockN{Charbel EL HACHEM}
\IEEEauthorblockA{\textit{FEMTO-ST Institute, CNRS}\\
\textit{Univ. Bourgogne Franche-Comte (UBFC)} \\
Belfort, France \\
charbel.el\_hachem@univ-fcomte.fr}
\and
\IEEEauthorblockN{Loïc PAINVIN}
\IEEEauthorblockA{\textit{R\&D Center}\\
\textit{Faurecia Clean Mobility}\\
Bavans, France\\
loic.painvin@faurecia.com}
}

\maketitle

\begin{abstract}
The welding seams visual inspection is still manually operated by humans in different companies, so the result of the test is still highly subjective and expensive. At present, the integration of deep learning methods for welds classification is a research focus in engineering applications. This work intends to apprehend and emphasize the contribution of deep learning model explainability to the improvement of welding seams classification accuracy and reliability, two of the various metrics affecting the production lines and cost in the automotive industry.
For this purpose, we implement a novel hybrid method that relies on combining the model prediction scores and visual explanation heatmap of the model in order to make a more accurate classification of welding seam defects and improve both its performance and its reliability.
The results show that the hybrid model performance is relatively above our target performance and helps to increase the accuracy by at least 18\%, which presents new perspectives to the developments of deep Learning explainability and interpretability.
\end{abstract}


\section{Introduction}
Welding is a manufacturing process consisting in joining two or more elements in a permanent way while ensuring continuity between these elements. The assembly is done either by heating, by pressure, or by the 2 combined. This process has an important role in productions lines, especially in the automotive industry, and is surely being influenced by the advent of digital technologies such as artificial intelligence, IoT, and intelligent manufacturing \cite{wang2020intelligent,egger2020augmented, yusof2020welding }.\\
Welding defect is an issue that affects Faurecia plants worldwide. Currently, there is a lot of research work being done on the automation of quality control \cite{el2021automation} and visual inspection of welding seams. Faurecia has developed automatic systems based on artificial intelligence to check the quality of welding seams in plants. Until now, these systems have been evaluated solely by their accuracy, the number of true positive and true negative. However, they have some shortcomings and errors in the classification of welding seam quality even when their accuracy is high \cite{el2021automation}.
Since these deep learning models work like black boxes, research is progressing on the explainability and interpretability of the results delivered by AIs \cite{london2019artificial}.
Nowadays, model interpretation is becoming a primary evaluation metric as well as its performance, and a good compromise between explainability and accuracy is more and more necessary in industrial applications.\\
In the study, the followed approach consists on:	\begin{enumerate}
	\item Taking advantage of a pre-trained model (MobileNet) for welding seams classification at Faurecia, tested by El Hachem et al. \cite{el2021seam}. 
		\item Implementing another deep learning model (ResNet-50) for comparison.
	\item Using visual explanation heatmap to analyse and interpret the results of these models and to seek improvement opportunities.
	\item Developing a hybrid approach taking into account model explainability to improve the overall accuracy.
	\end{enumerate}
The developed inspection system should be able to classify the defects with at least 97\% accuracy target, as requested by the client. \\This paper is organized as follows. Section II introduces the problem statement and motivation to set up this study. Section III presents an overview of the existing work done in the area of welding seam inspection and model explainability. In Section IV, the novel hybrid approach is introduced. In Section V, the implementation details are mentioned. In Section VI, the analysis of the experimental results is presented. The last section gives the concluding remarks.

\section{Problem Statement}
\subsection{Automated Visual Inspection}
The goal of automation is to use digital technologies in order to reduce the workload of the production operator, to improve the productivity and quality of the work performed, more specifically in the automotive industry. These systems take advantage of the vast evolution in the fields of artificial intelligence, computer vision, and robotics. To remain competitive, the companies use automated systems to make their plants more autonomous and reduce the human impact on the reliability of quality control. Automation also allows overcoming several other problems such as fatigue, security, data integrity, reliability, and availability. Hence the recent interest of companies to invest in this technological transition \cite{syed2020robotic}.

Faurecia produces exhaust systems that contain many welding zones. A leak test is done to check the quality of the welds after the welding operation is completed. Four welds are not checked by this test. That is why their automatic visual inspection is necessary to ensure the quality of the whole manufactured product.

\begin{figure}[hbt!]
    \centering
    \includegraphics[width=7cm]{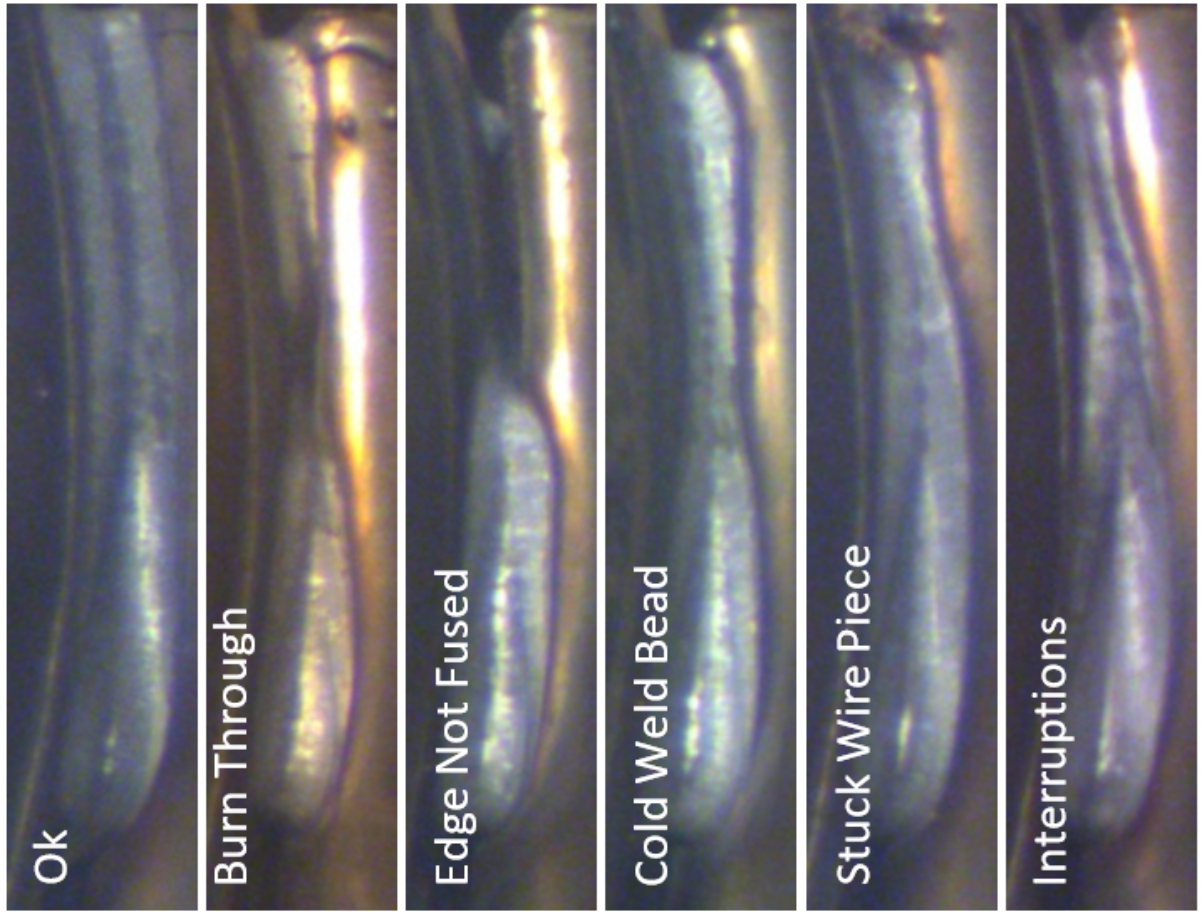}
     \caption{Reference weld (OK) and five possible defects (NOK)}
    \label{fig:weldsdef}
\end{figure}

    Figure \ref{fig:weldsdef} shows the different possible scenarios of the quality of the welds: only the first case is considered as OK.
\subsection{Actual Model Accuracy}
During all the tests done on different deep learning trained models, only a few were able to achieve the accuracy needed. The major source of low accuracy is that the two classes are not equally represented in the sample. Since weld defects are not potentially always present in production, photos from both the OK and NOK classes will not behave the same occurrence, resulting in the problem of unbalanced data which can be solved using resampling techniques \cite{mrozek2020efficient}. For this application, the problem was initially solved by using the data augmentation technique with various filters and transformations. Although, the accuracy can be further corrected by applying other modifications to the overall approach.

\subsection{Model Reliability}
As many deep learning applications are life or cost crucial, their accuracy is not the only evaluation metric that matters  \cite{alshemali2020improving}, especially in an industrial context. The model reliability is also considered as an important criterion to choose the best model. In fact, even if the accuracy target is achieved, the model might use a biased part of the image to classify it. This behavior represents a problem that is hard to detect and can generate a lot of damage since the model appears to perform well when in reality it is completely biased and its accuracy drops when faced with new data.\\

\begin{figure}[hbt!]
    \centering
    \includegraphics[width=6cm]{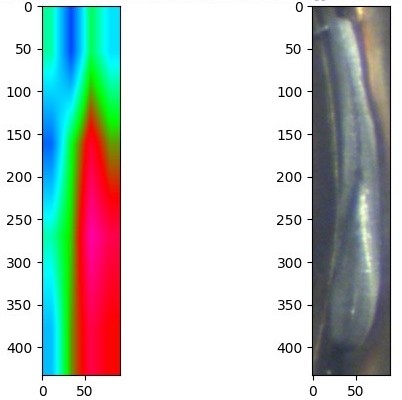}
     \caption{Model's activation heatmap of Misclassified NOK Weld}
    \label{fig:heatweld}
\end{figure}

Figure \ref{fig:heatweld} shows the model's activation heatmap of a misclassified weld. In reality, the weld is defective (NOK) but the model classifies it as of good quality (OK). It can clearly be seen that this is due to the false region on which the model relies to make its decision. Yet the model has good overall accuracy on the dataset. It is therefore necessary to study the reliability of the model in addition to its accuracy and to exploit the correlation between the activation heatmap results and the other scores of the model to make a more accurate decision.

\section{Related Works}
\subsection{Welding Seam Classification}
\begin{figure}[hbt!]
    \centering
    \includegraphics[width=9cm]{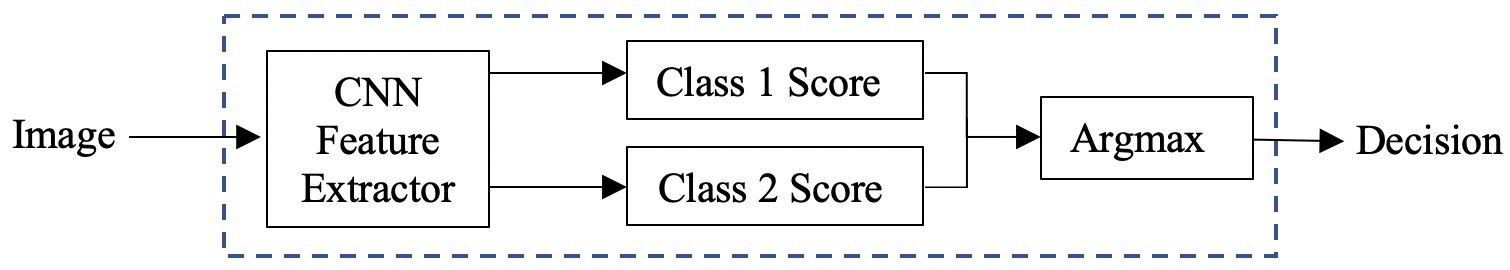}
     \caption{Old decision making approach}
    \label{fig:decision1}
\end{figure}

El Hachem et al. \cite{el2021seam} applied data augmentation to improve the accuracy of 4 welding seams classification. A deep learning model based on Mobilenet architecture was trained on a set of pictures of augmented weld defects. The image is initially fed to a convolutional neural network (feature extractor). After adjusting the weights, the network returns two scores for each image to classify it. The predicted class is the one with the highest prediction score. In general, the two prediction scores can either be class probabilities or class scores, which accept values larger than 1 (Fig.\ref{fig:decision1}).\\

The best results of this old approach are shown in table 1. 
Although the targeted accuracy (97\%) was achieved in some welds, it remains a challenge on other welds. Hence the need to look for other methods to overcome this problem.
\begin{table}[htbp]
\caption{Initial Accuracy based on standard approach}
\begin{center}
\begin{tabular}{l|l|}
\cline{2-2}
                             & Accuracy \\ \hline
\multicolumn{1}{|l|}{Weld 1} & 97,5\%   \\ \hline
\multicolumn{1}{|l|}{Weld 2} & 57,5\%   \\ \hline
\multicolumn{1}{|l|}{Weld 3} & 80,5\%   \\ \hline
\multicolumn{1}{|l|}{Weld 4} & 96,5\%   \\ \hline
\end{tabular}
\label{tab1}
\end{center}
\end{table}

Yang et al. \cite{yang2020weld} have constructed a novel deep learning framework to classify weld defects. With this new approach, the classification accuracy was improved by 3.18\% and 4.33\%. Even with all the work done, it is believed that there is still a large room for improving the classification performance over the generic DNN
models by taking into account other parameters.
\subsection{Visual Explanation Heatmap}

Knowing how to explain the reasons that led to a decision gives it more credibility and shows a strong capacity for reasoning and intelligence. The explainability of the AI model reinforces the credibility of the obtained results and also allows us to evaluate the reliability of the model and its behavior if a partial change of the data occurred. For example, in health care, the model verification and interpretation by medical experts is a necessity \cite{samek2017explainable}.
There are different methods that make the AI model explainable: Selvaraju et al. \cite{selvaraju2017grad} have first proposed the Grad-CAM method, a visualization technique able to tell which parts of a given image led the trained convolutional network to its final classification decision. This method makes it easy to debug the decision process of a CNN, especially in the case of misclassification. The result of this method is an activation heatmap indicating the parts of the image that have contributed the most to the final decision of the network.\\

\begin{figure}[hbt!]
    \centering
    \includegraphics[width=9cm]{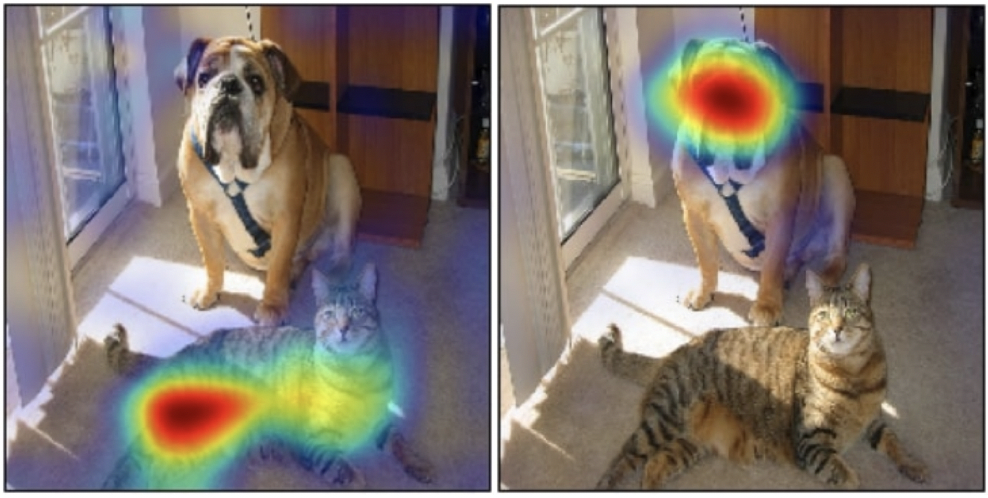}
     \caption{Example of Grad-CAM heatmap on Dog \& Cat Classification}
    \label{fig:dogcat}
\end{figure}

Figure \ref{fig:dogcat} shows the application of Grad-CAM heatmap on Dog \& Cat Classification. The model is accurate because it relies on the correct region of the image to classify it.\\
In the same context, Zhang et al. \cite{zhang2021grad} have applied model explainability to interpret the deep learning models trained to classify multiple sclerosis types in the brain using the Grad-CAM method. The experimental results showed that Grad-CAM gives the best heatmap localizing ability, and CNNs with a global average pooling layer and pre-trained weights had the best classification performance. 

\subsection{Hybrid Methods}

Ahlawat et al. \cite{ahlawat2020hybrid} have implemented a hybrid method using CNN and SVM classifier for handwritten digit recognition: CNN works as an automatic feature extractor and SVM works as a binary classifier. Their results showed that the hybrid approach achieved an accuracy of 99.28\% for the MNIST dataset.\\
Soumaya et al. \cite{soumaya2021detection} have tested a hybrid classification model using a genetic algorithm and SVM to detect Parkinson's disease. Their method attempts to give an accuracy of 80\% and 72.50\% using two kernels of SVM. The hybrid method seems to ensure the optimization of the classification system by minimizing the dimension of the features vector and maximizing the accuracy.\\
In the same context, Ahammad et al. \cite{ahammad2020hybrid} have suggested a new CNN-deep segmentation-based boosting classifier for spinal cord injury prediction. This method gives 10\% improvement on the classification rate.\\
Another hybrid method is developed by Liu et al. \cite{liu2018hybrid} for CO$_2$ welding defects detection by using CNN for primary features extraction and LSTM for feature fusion. The algorithm reaches 94\% of prediction accuracy.

\subsection{Frozen Graph Representation}

The automated visual systems shall be deployed on the plants. For this, it is preferable to have a trained model file as light as possible. The Tensorflow framework proposes the concept of frozen graph which is used to save what is necessary for the model while performing inference: the model is said to be frozen when keeping only the parts of it that matter to the deployment phase.  However, access to the frozen model weights becomes more difficult. Hence the need to develop a solution to save the model as a frozen graph without making it harder to extract or restore the data from its layers when performing a model explainability method.
A frozen graph in TensorFlow is saved as a file with the extension ".pb". It stands for protocol buffers, a language-neutral, platform-neutral extensible mechanism for serializing structured data as defined by its developer Google \cite{google}.

%%%% No need to "our" and "we"
%%%% Use passive mode
%%%% Use small sentences

%%% add correlation map curve 
%%% inverse columns and rows

\section{Methods}

\subsection{Convolutional Neural Networks}
Convolutional Neural Networks (CNN) have undergone a remarkable evolution with the development of new computational tools. These neural networks perform themselves all the tedious work of feature extraction and description: during the training phase, the classification error is minimized in order to optimize the parameters and the feature extraction. Moreover, the specific architecture of the network allows the extraction of features of different complexity, from the simplest to the most sophisticated in the image, hence the name "feature extractor". \\
CNN achieved very interesting results on different challenging datasets such as ImageNet \cite{krizhevsky2012imagenet}. Which shows its high performance in solving such vision problems.\\
For classification problems, the neural network calculates from the input image a score (or probability) for each class. The class attributed to the input object corresponds to the one with the highest score. The values of the layer weights are learned by backpropagation of the gradient \cite{hecht1992theory}: the parameters that minimize the regularized loss function are computed progressively, for each layer, starting from the end of the network. The optimization is done with stochastic gradient descent.\\

\subsection{Model Explainability}
\begin{figure}[hbt!]
    \centering
    \includegraphics[width=6cm]{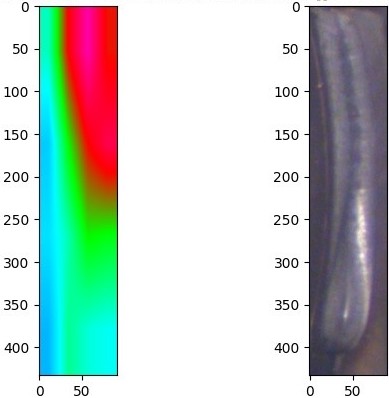}
     \caption{Model's Activation heatmap of a welding seam}
    \label{fig:gradcam1}
\end{figure}

The main approach of this work is based on the Grad-CAM method, introduced by Selvaraju  et  al. \cite{selvaraju2017grad}. The method assigns importance to each position in the last convolutional layer by computing the linear combination of activations, weighted by the corresponding output weights for the predicted class. The resulting class activation mapping is then resampled to the size of the input image.\\
 This technique will be used as a reliability test of the deep learning model by comparing the area with high color intensity to the real area of interest in the image. The model is more reliable if the two areas have closer values and positions. \\ 
The implemented algorithm generates an activation heatmap showing the region with the highest impact on the final decision (fig. \ref{fig:gradcam1}).

The second step is the calculation of the Red Color Ratio (RCR) in the heatmap. A K-nearest neighbor algorithm (KNN) is used to classify and identify precisely the pixels' closest color in the heatmap with K = 3 representing the RGB color system. In the end, the Red Color Ratio is calculated by dividing the number of pixels considered as red to the :\\ 
\begin{equation*}
    RCR = \frac{\text{Number of Red Cluster pixels}}{\text{Total number of pixels}} \times 100
\end{equation*}
By plotting the Red Color Ratio (RCR) in the heatmap as a function of the class scores, a high correlation is identified. This reinforces the idea that weld defects are often labeled by a limited feature zone, compared to non-defective welds. RCR is considered as a reliability score of the model prediction.
\begin{figure}[hbt!]
    \centering
    \includegraphics[width=9cm]{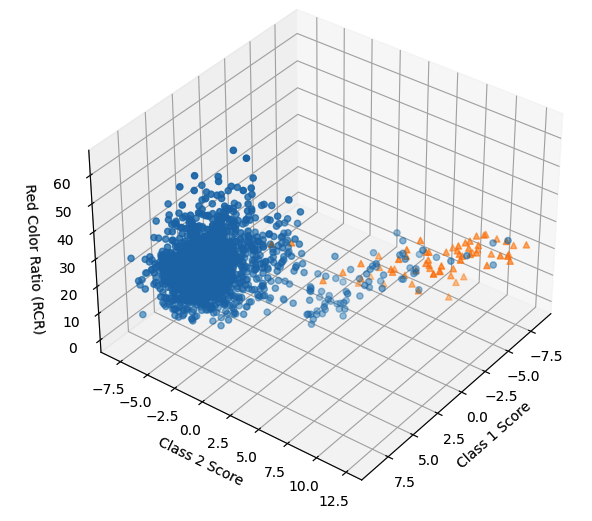}
     \caption{Distribution of images per class scores and RCR}
    \label{fig:dist}
\end{figure}

The figure \ref{fig:dist} shows the distribution of the weld 1 images per class scores and RCR. Each blue dot is a non-defective weld in reality (OK Class), and the orange triangle is a defective weld (NOK Class). This identified correlation was the major motivation to investigate this direction (fig. \ref{fig:dist}). The dataset is now more separable than when represented by only two scores. 

\begin{figure}[hbt!]
    \centering
    \includegraphics[width=9cm]{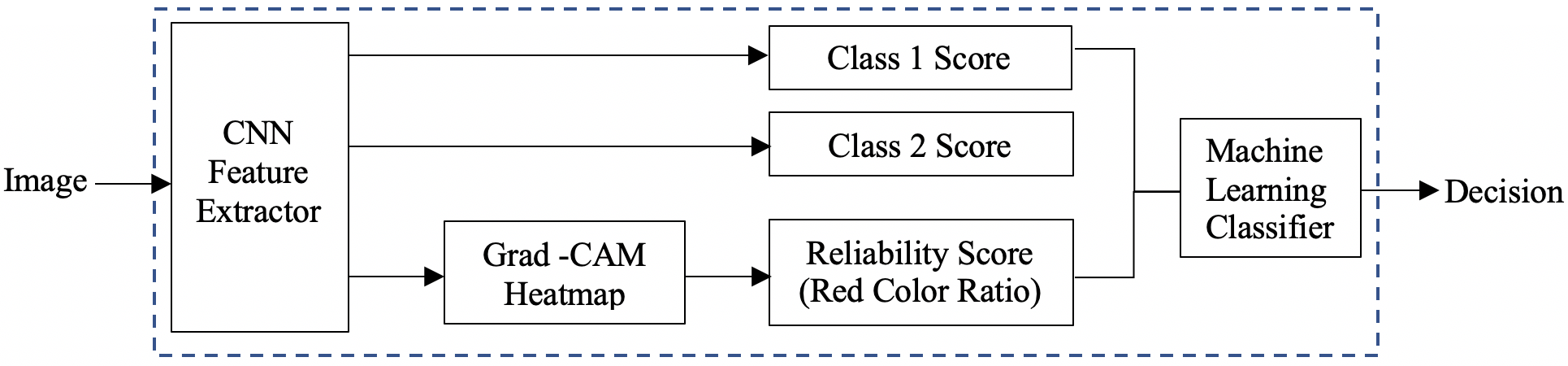}
     \caption{Novel decision making approach}
    \label{fig:gradcam2}
\end{figure}

The use of 3 scoring parameters (Class 1 Score, Class 2 Score, Red Color Ratio) may help to get more accurate thresholding for the classification of these points. This is the main reason behind the development of this new model explainability-based approach (fig. \ref{fig:gradcam2}).\\

\section{Implementation Details}

In this study, two deep learning models were used: the MobileNet architecture that has been trained on the same welding seams dataset by El Hachem et al. and a ResNet-50 model that is trained during this study for comparison.

\subsection{Experimental Environement}
The experimental environment is powered by Intel i5 CPU, 2.30 GHz with 64-bit, Windows 10 system and 8 GB memory. The software programming environment is Python. It uses both Keras and Tensorflow as backend.\\

\subsubsection{Feature Extractors}
For the ResNet-50 model, Adam is selected as optimizer of the CNN with a learning rate set to $10^{-5}$. A pre-trained version of the model weights trained on the ImageNet database has been loaded. The input of the CNN is represented by the vector (batch size, height, width, depth). The batch size (equal to 32) defines the number of samples that will be propagated through the whole network. The other parameters represent the height, width, and depth of the image. The input has the following values (32, 128, 128, 3). 
For the Mobilenet model, Rmsprop is selected as the optimizer of the CNN. The chosen learning rate decay type is exponential starting with a value of 0.01 and ending with a value equal to 0.0001. The model has been trained with 9000 epochs.\\
\subsubsection{Machine Learning Classifiers}
The XGboost Classifier is a gradient boosting algorithm that offers a very large panel of hyperparameters, it is thus possible, thanks to this diversity of parameters, to have total control over the implementation of Gradient Boosting \cite{DBLP:journals/corr/ChenG16}. The chosen booster is "gbtree" which uses a tree-based boosting. The step size shrinkage is set to 0,5 and the used sampling method is the uniform selection.

The decision tree uses the Gini function to measure the quality of the split. The max depth is set to the default value so that the tree nodes are expanded until all leaves are pure or contain less than 2 elements.

On the other hand, Support Vectors Machines Classifier (SVM)
are a family of machine learning algorithms that solve classification and regression problems. They are known for their strong theoretical guarantees and their great flexibility.
The idea of the Support Vector Machine is to project the data into a higher-dimensional space and to make them separable. SVMs become a universal approximator \cite{wang2004rbf}: with enough data, the algorithm can always find the best possible boundary to separate two classes.\\
There are several formulations of SVM, each formulation makes a specific transformation of the input data and has a specific shape of the decision boundaries, these formulations are called kernels. In this study, two SVM kernels are used:
\begin{itemize}
	\item Linear Kernel: This is the case of a linear classifier, without space change. The data are assumed to be linearly separable, i.e. there is a hyperplane that separates the data without error. This is not quite the case observed on the correlation curve above. However, it is still worthwhile to see the effect of the separation by this kernel on the final accuracy.
	
		\item 5-Polynomial Kernel: This is the case where we change the space by applying a polynomial transformation (in this case a polynomial of degree 5). The choice of the decision boundaries will be based on the new space.
\end{itemize}

In total, four different machine learning classifiers are tested for the improvemeent of the decision making part on the two CNN feature extractors. 

\section{Experimental Results}

Figure 8 shows the results obtained on the weld 1, weld 2, weld 3, and weld 4 using MobileNet and ResNet-50 models based on the old approach. The target is only reached with the weld 1 using MobileNet architecture.

\vspace{0.5cm}

\begin{figure}[h!]
  \begin{center}
    \begin{tikzpicture}
      \begin{axis}[
          width=\linewidth, % Scale the plot to \linewidth
xlabel=Weld Index,
  ylabel=Accuracy (\%),
          legend style={at={(0.8,0.2)},anchor=north}, % Put the legend below the plot
          x tick label style={rotate=90,anchor=east} % Display labels sideways
        ]
\addplot table [x=P, y=$MobileNet$]{data.dat};
\addlegendentry{$MobileNet$}
\addplot table [x=P, y= $ResNet$]{data.dat};
\addlegendentry{$ResNet$}
        
        \legend{MobileNet, ResNet-50}
      \end{axis}
    \end{tikzpicture}
    \caption{Comparison between MobileNet \& ResNet - Old Approach}
  \end{center}
\end{figure}
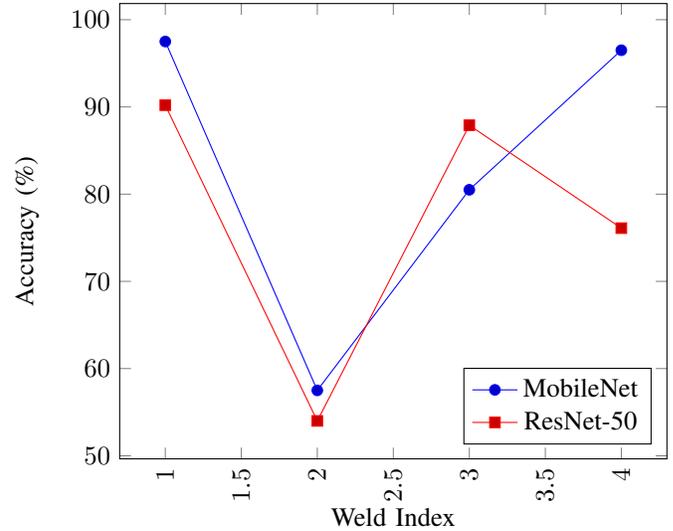

\begin{table}[htbp]
\caption{Mobilenet Model Accuracy - New Approach}
\begin{center}
\begin{tabular}{|c|c|c|c|c|}
\hline
\textbf{Welding}&\multicolumn{4}{|c|}{\textbf{Accuracy per Classifier}} \\
\cline{2-5} 
\textbf{Seams} & \textbf{\textit{XGboost}}& \textbf{\textit{Decision Tree}}& \textbf{\textit{SVM Linear}}& \textbf{\textit{SVM Poly5}} \\
\hline
Weld 1& 98,4\%& 98,4\%& \textbf{99,3\%}& 98,9\% \\
Weld 2& \textbf{99,3\%}& 98,1\% & 98,8\%&98,8\%  \\
Weld 3&\textbf{98,7\%} & 97,2\%&  98,1\%&\textbf{98,7\%} \\
Weld 4& \textbf{98,1\%}&97,9\% & \textbf{98,1\%}& \textbf{98,1\%}\\
\hline
\end{tabular}
\label{tab1}
\end{center}
\end{table}

Table II proves that adding a statistical machine learning classifier after the feature extractor and class activation heatmap seems to increase the overall accuracy of the model for all of the four welds. The target accuracy is acheived by the four machine learning classifiers, with a tiny advantage to XGboost classifier.

On the other hand, Table III explains how hybrid ResNet-50 behaves when the network is followed by a machine learning classifier. The accuracy is better than the standard approach but still smaller than Mobilenet.

\begin{table}[htbp]
\caption{ResNet-50 Model Accuracy - New Approach}
\begin{center}
\begin{tabular}{|c|c|c|c|c|}
\hline
\textbf{Welding}&\multicolumn{4}{|c|}{\textbf{Accuracy per Classifier}} \\
\cline{2-5} 
\textbf{Seams} & \textbf{\textit{XGboost}}& \textbf{\textit{Decision Tree}}& \textbf{\textit{SVM Linear}}& \textbf{\textit{SVM Poly5}} \\
\hline
Weld 1& \textbf{97,1}\%& 95,1\%& 96,1\%& 95,3\% \\
Weld 2& \textbf{98,4\%}& 95,1\% & 96,8\%&98,8\%  \\
Weld 3&97,3\% & 92,8\%&  96,7\%&\textbf{98,1\%} \\
Weld 4& \textbf{97,9\%}&95,9\% & 97,2\%& \textbf{97,9\%}\\
\hline
\end{tabular}
\label{tab1}
\end{center}
\end{table}

\section{Conclusion \& Future directions}

In this paper, a hybrid approach of CNN-ML Classifier is proposed for welds defects classification. This approach adds a new reliability score calculated using the model explainability methods, more specifically Grad-CAM heatmap. This new score makes the dataset more separable.
The hybrid approach proposed in this paper seems to perform very well on weld defects classification. The highest accuracy improvement was by +71\% for weld 2 using MobileNet-XGboost classifier and by +18\% for the weld 3 using MobileNet-SVM Poly5 Kernel.
MobileNet architecture performs better than ResNet-50 in extracting the feature from the dataset images. For the classifier part, XGBoost achieves higher accuracy and gives promising results comparing to SVM Linear kernel, Poly5 kernel, and Decision Tree at last.\\
This work presents new perspectives on the developments of new model-driven optimization methods to improve the accuracy of vision systems. In future work, it can be extended to other classification problems with highly unbalanced datasets.

\section*{Acknowledgment}

This work was done as a part of a CIFRE (N 2018/1029) project with Faurecia, funded by the Ministry of Higher Education and Research of France, managed by the Association Nationale de la Recherche et de la Technologie (ANRT) and was partially supported by the EIPHI Graduate School (contract ”ANR-17-EURE- 0002”).

\bibliographystyle{unsrt}
\bibliography{my}

\begin{appendices}

\end{appendices}

\end{document}